%% file: stability-gap.tex
\documentclass[10pt,twocolumn,letterpaper]{article}

\usepackage{cvpr}    

\input{preamble}

\definecolor{cvprblue}{rgb}{0.21,0.49,0.74}
\usepackage[pagebackref,breaklinks,colorlinks,citecolor=cvprblue]{hyperref}
\usepackage{comment}
\usepackage{soul}
\newcommand{\minisection}[1]{\vspace{0.005in} \noindent {\bf #1}}

\title{The Expanding Scope of the Stability Gap: Unveiling its Presence in Joint Incremental Learning of Homogeneous Tasks}
\author{Sandesh Kamath\textsuperscript{1,2} \enspace\enspace Albin Soutif-Cormerais \textsuperscript{1,2} \enspace\enspace 
Joost van de Weijer \textsuperscript{1,2} \enspace\enspace 
Bogdan Raducanu \textsuperscript{1,2}  \\ 
\textsuperscript{1}Department of Computer Science, Universitat Autònoma de Barcelona \\
\textsuperscript{2}Computer Vision Center, Barcelona\\
{\tt\small \{skamath, albin, joost, bogdan\}@cvc.uab.es}
}

\begin{document}
\maketitle

\begin{abstract}
Recent research identified a temporary performance drop on previously learned tasks when transitioning to a new one. 
This drop is called the stability gap and has great consequences for continual learning: it complicates the direct employment of continually learning since the worse-case performance at task-boundaries is dramatic, it limits its potential as an energy-efficient training paradigm, and finally, the stability drop could result in a reduced final performance of the algorithm. In this paper, we show that the stability gap also occurs when applying joint incremental training of homogeneous tasks. In this scenario, the learner continues training on the same data distribution and has access to all data from previous tasks. In addition, we show that in this scenario, there exists a low-loss linear path to the next minima, but that SGD optimization does not choose this path. We perform further analysis including a finer batch-wise analysis which could provide insights towards potential solution directions.
\end{abstract}

\section{Introduction}
Deep neural networks demonstrate remarkable performance across numerous machine-learning tasks. Nevertheless, when trained on non-IID streaming data these networks struggle to accumulate knowledge, and tend to forget previously acquired knowledge. Continual learning develops theory and methods to address this problem~\cite{delange2021continual, masana2022class}. It aims to develop algorithms that prevent \textit{catastrophic forgetting}~\cite{mccloskey1989catastrophic} and achieve a more favorable trade-off between stability and plasticity~\cite{mermillod2013stability} while learning on a data stream.

A typical test setting that continual learning considers is learning from a sequence of tasks (each task with another data distribution)~\cite{delange2021continual}. Usually, continual learning method performance is evaluated at the end of each of the tasks. Recently, researchers~\cite{caccia2022new,lange2023continual} have observed an interesting phenomenon that went unnoticed in this standard evaluation setup: at the start of training a new task, the performance of previous tasks drastically drops, and only slowly recovers during the subsequent training of the new task. De Lange et al.~\cite{lange2023continual} coined the term \textit{stability gap} for this phenomenon. This observation should be taken into account for the application of continual learning systems (especially in safety-critical contexts) since it significantly lowers the worst-case performance of these algorithms. Furthermore, it can potentially worsen the final accuracy of the learner, since it might not recover totally from the knowledge loss incurred during the stability gap. Addressing the stability gap is therefore of utmost importance~\cite{harun2023overcoming,albin2023online}.

The underlying mechanism responsible for the stability gap remains the subject of lively scientific debate, with no clear explanation available yet. Originally, Caccia et al.~\cite{caccia2022new} hypothesized that the cause for the stability gap is because old class prototypes receive a large gradient from closely lying new class prototypes. However, this hypothesis could not fully explain the phenomenon, because the stability gap had also been observed in domain incremental learning (where the set of classes remains the same)~\cite{lange2023continual}. A possible remaining explanation is the following. When optimizing on new data, the objective is to minimize the loss on both the available new data and unavailable old data. The loss on the unavailable previous data is then approximated with various continual learning strategies, such as regularization~\cite{Kirkpatrick2016OvercomingCF, Li2016LearningWF} and data rehearsal~\cite{rebuffi2017icarl, chaudhry2019tiny}. An explanation for the stability gap could be the failure to approximate this ideal joint loss on previous and current task data. Surprisingly, a recent paper~\cite{hess2023complementary}, showed that even in the case of joint incremental training the stability gap occurred (in this case, we do have access to both old and new task data and can minimize the joint loss on both old and new data). They, therefore, came to the important realization that we should not only focus on \textit{what} to optimize but more importantly on \textit{how} to optimize our objective. 

\begin{table}[tb]
    \centering
    \resizebox{0.45\textwidth}{!}{
    \begin{tabular}{llll}
    \hline
     {Paper}  & \multicolumn{1}{c}{Type} & \multicolumn{2}{c}{Tasks}  \\
    \hline
         \citet{caccia2022new}  & CI: $c_{1} \neq  c_{2}$ & disjoint & heterogeneous \\
         \citet{lange2023continual}  &  DI: $c_{1} = c_{2}$  & disjoint & heterogeneous\\
         \citet{hess2023complementary}  &  CI: $c_{1} \neq  c_{2}$ &  joint incr. & heterogeneous \\
          \hline
Ours &  DI: $c_{1} = c_{2}$ & joint incr. & homogeneous\\
    \hline
    \end{tabular}}
    \caption{Summary of the expanding scope of the stability gap: from heterogeneous to homogeneous tasks. } 
    \label{tab:stability-gap}
\end{table}

\begin{figure*}[!ht]
    \begin{center}
    \includegraphics[width=0.48\textwidth]{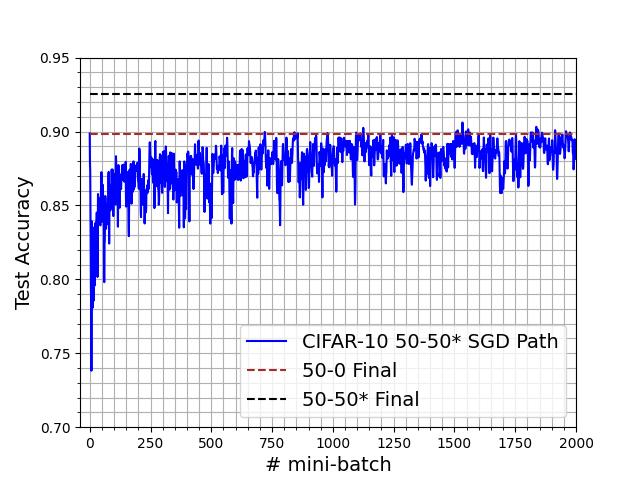}       
    \includegraphics[width=0.48\textwidth]{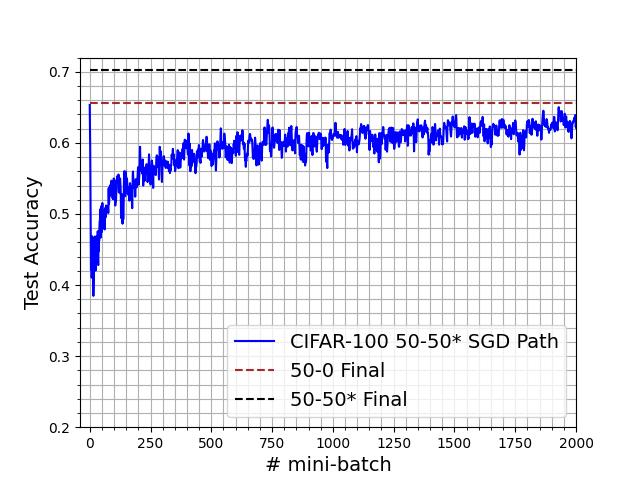}           
    \end{center}
    \caption{Occurrence of the stability gap in joint incremental learning with homogeneous tasks in the $50\text{-}50^*$ setting on (left) CIFAR-10 and (right) CIFAR-100 datasets on a ResNet-18 model. This plot starts after training with task A, and the x-axis represents the number of iterations of training on task B.}
    \label{fig:concept-diagram}
\end{figure*}

Hess et al.~\cite{hess2023complementary} made their important observation when learning on heterogeneous tasks, referring to the fact that each task is drawn from a different distribution. In this paper, we show that the stability gap is even present in the case of joint incremental learning on homogeneous tasks (where each task is drawn from the same distribution). This result in presented in Figure \ref{fig:concept-diagram}. So, even in the case that both tasks have the same distribution, SGD optimization does not succeed in going to the `nearby' optimal position without derailing through a high-loss region. The only difference between the new and old data is that the network has seen the old data (typically for 100 epochs here) and has not yet seen the newly arriving data. We think that this further confirms the fundamental nature of the stability gap in continual learning: it even occurs in the most simple continual learning setting when training from an increasing amount of data drawn from the same distribution. 
The main contributions of this work are: 
\begin{enumerate}
    \item We show that the stability gap also occurs during joint incremental learning from homogeneous tasks, arguably the least challenging continual learning setting.
    \item We show that there exists a linear low-loss path to the optimal loss, but that SGD is not following this path (this was hypothesized in~\cite{hess2023complementary} but was not demonstrated).
    \item We perform an analysis at mini-batch level, and discover that the gradient just after the task boundary successfully decreases the mini-batch loss but results in an overall loss increase on the test set. Addressing this might potentially lead to a solution to the stability gap problem.  
\end{enumerate}
This manuscript does not provide a new possible explanation for the stability gap. We think the observation that it occurs even for joint incremental learning of homogeneous tasks is relevant. Our results, confirm those of Hess et al.~\cite{hess2023complementary} and we agree with them that the focus should shift to how to optimize rather than what to optimize. 

\section{Stability Gap Analysis} \label{sec:stability-gap}

\subsection{Experimental setup} \label{subsec:exp-setup}
\minisection{Datasets:}  We use the standard benchmark train-test split for all the datasets used in this work, that is publicly available. 
CIFAR-10 dataset consists of $60,000$ images of $32 \times 32$ size, divided into $10$ classes: $50,000$ used for training and $10,000$ for testing. CIFAR-100 dataset consists of $60,000$ images of $32 \times 32$ size, divided into $100$ classes: $50,000$ used for training and $10,000$ for testing. 

\minisection{Architectures:} We consider two convolutional network architectures,  VGG-16 \citep{simonyan2015deep} and ResNet-18 \citep{he2015deep} for our study.

\minisection{Training Setup:} Our code base uses the pytorch library. For training we use the SGD optimizer with hyperparameters: learning rate (lr) of 0.01, momentum (m) of 0.9, batch size (bs) of 64.

\minisection{Notation:} 
In this work, we mainly study the two-task setting. All results reported will be in the \textit{homogeneous task setting}, where the various tasks are drawn from the same distribution. 
We use the notation of $A\textrm{-} B$ to indicate task A will contain A\% of the data and task B will contain B\% of the data from the original training dataset. We will use the notation $A\textrm{-} B^*$ to identify the \textit{joint incremental learning} setting. In this case when training task B, the algorithm has access to all the data of task A. In practice, for this setting for task B, we just combine the data of both tasks, and continue training on the combined dataset. Note, that the data of task A and B in our paper are disjoint data sets and do not contain the same data samples.

\minisection{Note on plots: }  
Most plots in this paper are with a warm-started model. This means a model trained on task A with the data as prescribed in the setting was used to continue training on task B. The starting point of the x-axis is then the iterations directly after the task-switch. This was done to better study the effect of the actual stability gap. Note, that we do not show the end of training on task B. 

\subsection{Stability Gap in Joint Incremental Learning of Homogeneous Tasks} \label{subsec:stability-gap-overlap}

To establish the occurrence of the stability gap in the setting of joint incremental learning of homogeneous tasks, we study the 50-50* setting. This setting divides the training set into two equally sized tasks, A and B. Both tasks are drawn from the same distribution. The test accuracy is provided for two datasets in Figure \ref{fig:concept-diagram}. We can observe that even in for this case, there is a clear stability gap. The performance drops from 0.89 to 0.74 on CIFAR-10 and from 0.65 to 0.38 on CIFAR-100. We posit a larger gap on CIFAR-100 to be related to the smaller number of samples per class. Note that for both these graphs performance has not returned to its task A level consistently even after the 2000 iterations showing the long-lasting impact of the stability gap. 
After continued training for around 3500-4500 iterations the models start to achieve more consistently a performance above 0.89 and 0.65, respectively. 

In Table~\ref{tab:stability-gap} we provide a summary of the main papers on the stability gap. The stability gap has been observed in increasingly general settings. Here, we show that it is also observed for joint incremental training of homogeneous tasks, which is arguably the most simple continual learning setting. This observation is relevant since it discards explanations for the stability gap which are based on characteristics that are not present (e.g. it cannot be uniquely explained by the presence of disjoint tasks or heterogeneous distributions).

\subsection{Linear Mode Connectivity}  \label{subsec:analysis-problem}

\citet{garipov2018loss} were the first to study the mode connectivity properties of neural networks weights by connecting two independent minima obtained through differently seeded optimization processes using a simple curved path of low loss. \citet{frankle2020linear} later showed that a simpler kind of path naturally emerges early in training. They observed that models that are trained from a warm-started model version on the same dataset but with different SGD-noise lead to two checkpoints that are connected by a linear path of low loss. \citet{mirzadeh2020linear} later extended that property to optima of multitask models trained on incrementally larger datasets. \citet{hess2023complementary} hypothesized that there exists a low-loss path between the optima when doing joint incremental learning of heterogeneous tasks. However, they do not demonstrate this in their paper. 
In this article, we investigate whether it is the case that training on incremental homogeneous tasks leads to linearly connected optimas or not (and we verify this). To do so, we  take the initial checkpoint with weights $\theta_1$ and final checkpoint with weights $\theta_2$ and interpolate between the two by taking $\theta_{\lambda} = \lambda\theta_1 + (1 - \lambda\theta_2)$ with $\lambda \in [0, 1]$. We later compute and report the test accuracy of each $\theta_{\lambda}$ to determine if the linear path is of low loss.

\begin{figure}[tb]
    \centering
    \includegraphics[trim=0 0 20 20, clip, width=0.49\linewidth]{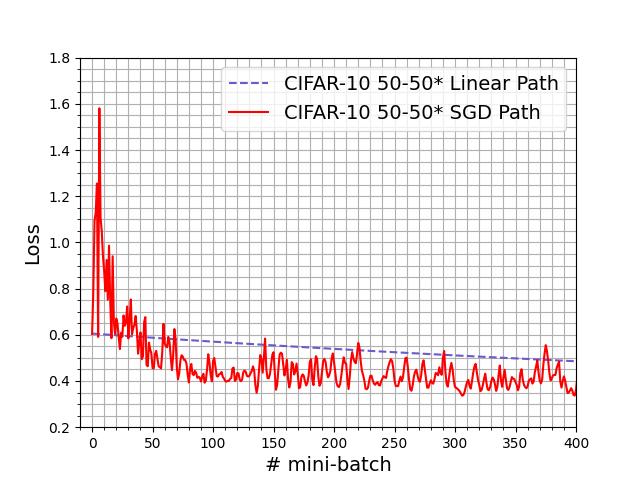}
    \includegraphics[trim=0 0 20 20, clip, width=0.49\linewidth]{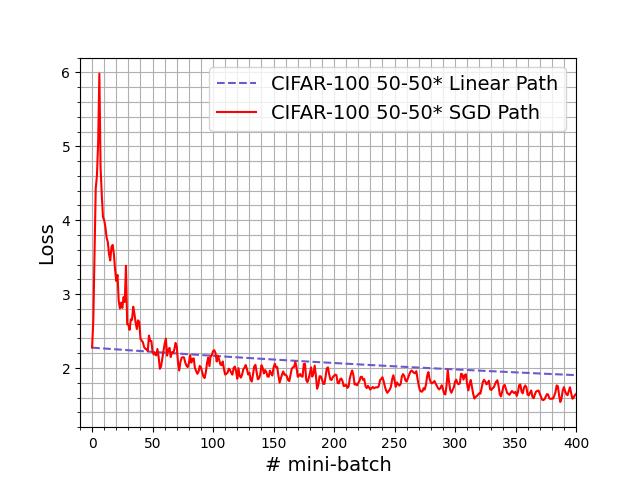}
    \caption{In the 50-50* setting, we present the loss path with SGD and the linear connectivity loss path between the warm-start and final models using with ResNet-18 model on (left) CIFAR-10, (right) CIFAR-100 dataset. In order to observe the stability gap, we zoom in on the first few iterations of the new task.}
    \label{fig:linear-mode-connectivity}
\end{figure}

Figure~\ref{fig:linear-mode-connectivity} compares the loss of the models obtained by linearly interpolating between the initial and final model to the ones of the model checkpoints along the SGD optimization trajectory. Unsurprisingly, the path taken by SGD during optimization is not linear. More surprisingly, it goes through areas of higher loss especially during the initial period that corresponds to the \textit{stability gap}, while the linear path between the initial and final model is of low loss. The linear path results confirm that a low-loss path exists between the minima achieved after training task A, and the minima after training task B. Surprisingly, SGD does not take this path and instead passes through a high-loss area before converging towards the minima which is optimal for task A and B data. We have shown here the first few iterations after the task-switch. Refer to the supplementary for plots showing the entire path based on 5 epochs of training with task B.

\begin{figure}[]
    \centering
    \includegraphics[width=1.0\linewidth]{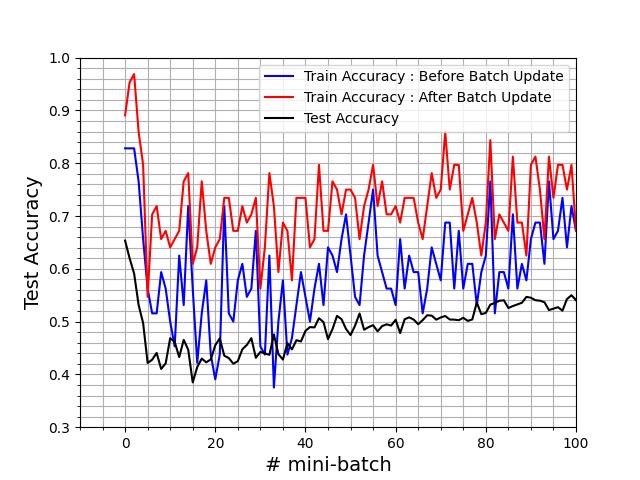}
    \caption{Using CIFAR-100 with ResNet-18, we present the finer analysis of the \textbf{local} improvement obtained at the batch level by observing the train accuracy per batch before (\textcolor{blue}{blue} line) and after (\textcolor{red}{red} line) SGD update is applied for the batch in the 50-50* setting. The black line is the corresponding test accuracy.}
    \label{fig:training-batchwise-results}
\end{figure}

\minisection{Per mini-batch loss analysis.} In Figure~\ref{fig:training-batchwise-results}, we observe with a microscope the learning of the model per mini-batch. In this plot we show the training batch accuracy for the current mini-batch before (\textcolor{blue}{blue} line) and after (\textcolor{red}{red} line) the SGD update. We observe that the SGD update results in a loss decrease(or accuracy increase) for the particular mini-batch (the \textcolor{blue}{blue} line is below the \textcolor{red}{red} line). 
However, when we look at the test accuracy (black line), we see that even though initial steps lead to a lower loss on the mini-batch, they do not result in better test performance. The black line goes down in the initial iterations. This means that the SGD update moves the network parameters away from the optimal path. 

\subsection{Additional Analysis}
Here we verify if the stability gap also occurs for several other settings. 

\begin{figure}[]
    \centering
    \includegraphics[trim=0 0 20 20, clip, width=0.49\linewidth]{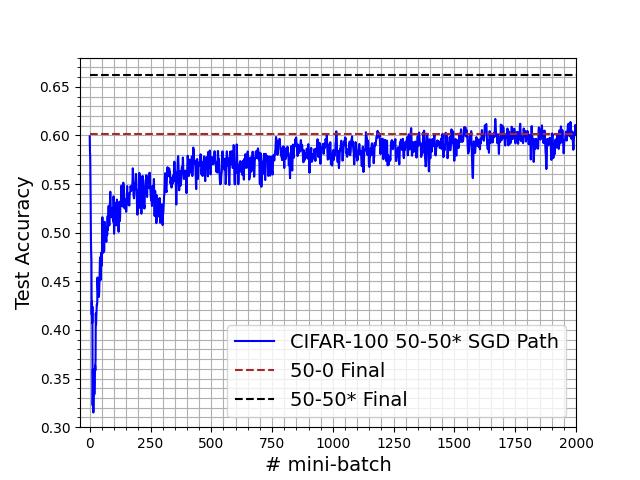}
    \includegraphics[trim=0 0 20 20, clip, width=0.49\linewidth]{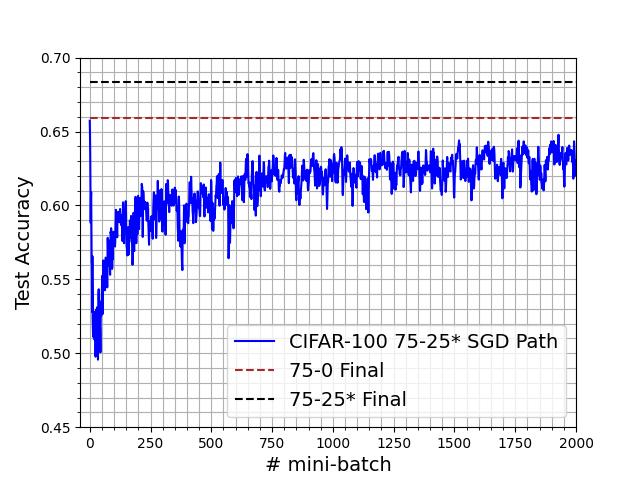}
    \caption{Using CIFAR-100 with VGG-16, stability gap in (left) 50-50* (right) 75-25* setting.}
    \label{fig:training-75-100-batchwise-results-vgg16}
\end{figure}

\minisection{Stability gap using other architectures.} While we present a detailed study of the stability gap on ResNet-18 architecture, in Figure \ref{fig:training-75-100-batchwise-results-vgg16} we show this phenomenon is not restricted to a specific architecture by using another well-known VGG-16 architecture on the CIFAR-100 dataset.

\begin{figure}[]
    \centering
    \includegraphics[trim=0 0 20 20, clip, width=0.49\linewidth]{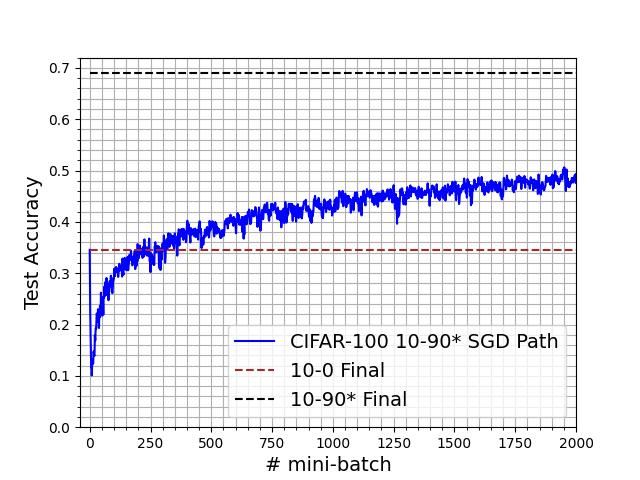}
    \includegraphics[trim=0 0 20 20, clip, width=0.49\linewidth]{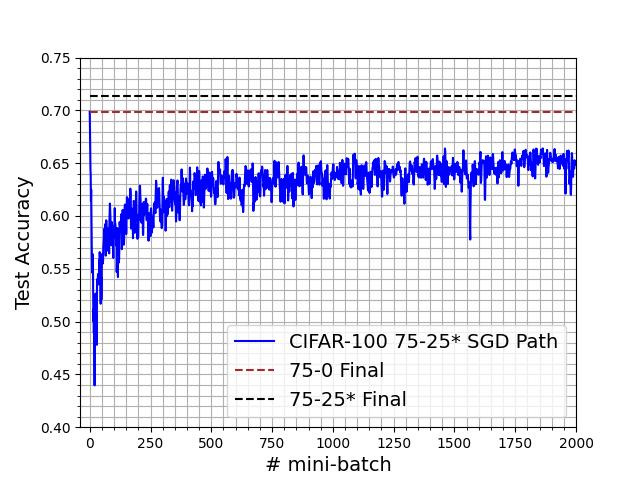}
    \caption{Using CIFAR-100 with ResNet-18, stability gap in (left) 10-90* (right) 75-25* setting. We can see that the stability gap increases for a smaller-sized first task.}
    \label{fig:training-10-75-100-batchwise-results}
\end{figure}

\minisection{Stability gap in other settings.} In Section \ref{subsec:stability-gap-overlap}, we mainly considered the 50-50* setting which is the joint incremental training with homogeneous task. 
Here, we look at the stability gap with different first task size and include results for the setting 10-90* and 75-25* in Figure \ref{fig:training-10-75-100-batchwise-results}. We  observe that the gap is larger when starting from a smaller first task. 

\begin{figure}[]
    \centering
    \includegraphics[trim=20 0 20 20, clip, width=0.49\linewidth]{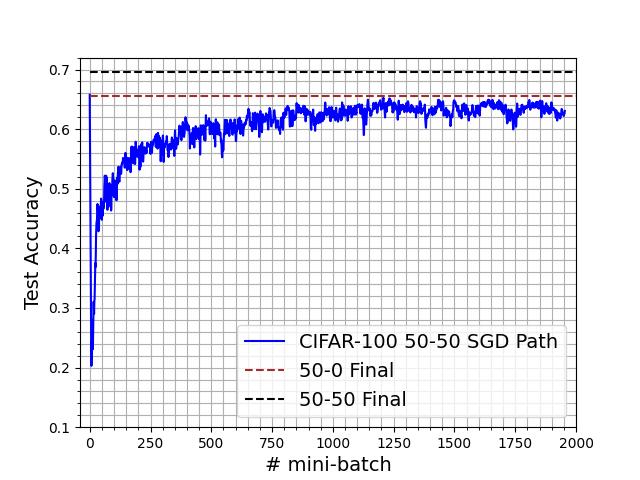}
    \includegraphics[trim=20 0 20 20, clip, width=0.49\linewidth]{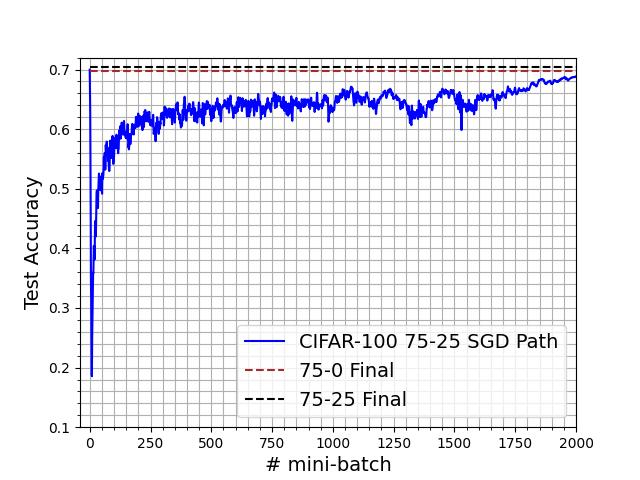}
    \caption{Using CIFAR-100 with ResNet-18, stability gap in (left) 50-50, (right) 75-25 setting. We can see that the stability gap increases when comparing (left) with the 50-50* setting in Fig.~\ref{fig:concept-diagram}(right) and (right) with the 75-25* setting in Fig.~\ref{fig:training-10-75-100-batchwise-results}(right).}
    \label{fig:training-50-75-batchwise-results}
\end{figure}

In addition, we conduct experiments with the splits 50-50 and 75-25 which is equal to incremental training with new data from the same distribution (without access to all previous data). We observe in Figure \ref{fig:training-50-75-batchwise-results} that the stability gap occurs in this setting too and is more pronounced than the corresponding 50-50* and 75*-25* setting studied before. The gap is larger from 0.65 to 0.20 and 0.70 to 0.23 as against 0.65 to 0.38 and 0.70 to 0.44, respectively.

\section{Conclusions}
In this article, we present compelling insights into the stability gap phenomenon. In particular, we show that it also manifests when applying joint incremental training on a sequence of homogeneous tasks, which is often considered the simplest scenario for continual learning. Through experimental evidence, we demonstrate that while the loss along the SGD path displays a stability gap, this discrepancy is not mirrored in the loss along the linear trajectory between checkpoints. An analysis at the mini-batch level showed that the gradient computed on the initial mini-batches (after the task-switch) does reduce the loss for each mini-batch but it results in an increased loss on the test data. 
We also observe that in the incremental learning with homogeneous tasks, when we remove rehearsal (going 50-50* to 50-50), the stability gap increases.
In further research, we will explore this direction to possibly discover the cause of the stability gap and possible remedies.

\paragraph{Acknowledgement.} 
We acknowledge projects TED2021-132513B-I00 and PID2022-143257NB-I00 funded by MCIN/AEI/10.13039/501100011033, by European Union NextGenerationEU/PRTR, by ERDF A Way of Making Europe, and by Generalitat de Catalunya CERCA Program.

{
    \bibliographystyle{ieeenat_fullname}
    \bibliography{stability-gap}
}

\clearpage
\maketitlesupplementary

We have included more detailed empirical results in the supplementary material.

\begin{figure}[!ht]
    \centering
    \includegraphics[trim=20 0 20 20, clip, width=0.8\linewidth]{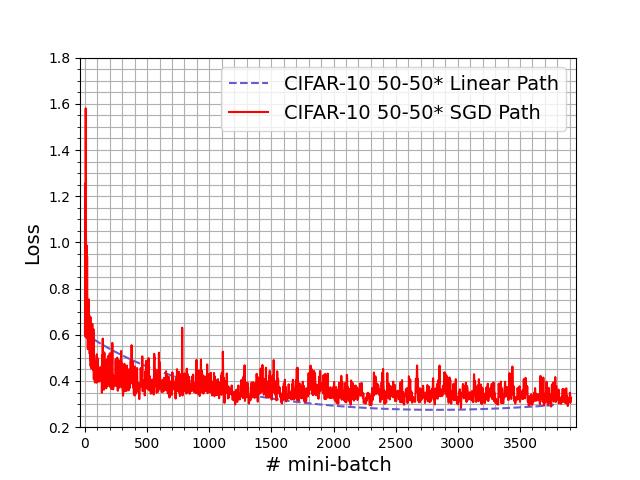}
    \caption{In the 50-50* setting, we present the complete loss path with SGD and linear connectivity between the warm-start and final models after 5 epochs of training using $\lambda$ changed in steps of 0.01 on CIFAR-10 dataset with ResNet-18 model. In order to observe the stability gap, we had zoomed on the first 400 iterations in the main paper.}
    \label{suppl:fig:linear-mode-connectivity-cifar10}
\end{figure}

\begin{figure}[!ht]
    \centering
    \includegraphics[trim=20 0 20 20, clip, width=0.8\linewidth]{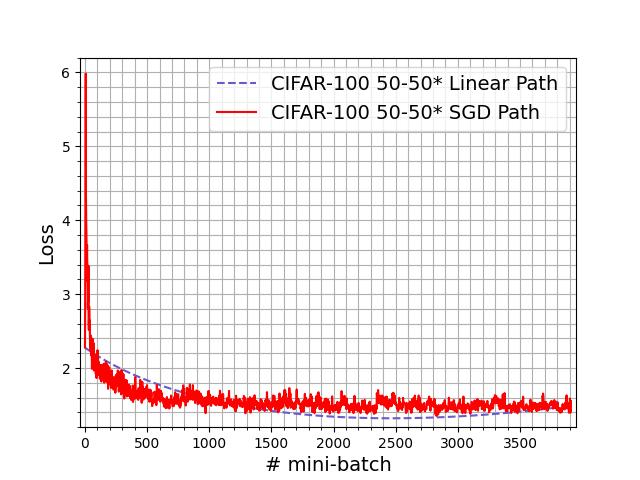}
    \caption{In the 50-50* setting, we present the complete loss path with SGD and linear connectivity between the warm-start and final models after 5 epochs of training using $\lambda$ changed in steps of 0.01 on CIFAR-100 dataset with ResNet-18 model. In order to observe the stability gap, we had zoomed on the first 400 iterations in the main paper.}
    \label{suppl:fig:linear-mode-connectivity-cifar100}
\end{figure}

\minisection{Linear Mode Connectivity}  \label{app:subsec:analysis-problem}
In Figure \ref{suppl:fig:linear-mode-connectivity-cifar10} and  Figure \ref{suppl:fig:linear-mode-connectivity-cifar100} we show the full loss trajectory of the second task training for 5 epochs for both the linear path and the SGD path for CIFAR-10 and CIFAR-100 datasets, respectively. 

\end{document}

%% file: preamble.tex
\usepackage[dvipsnames]{xcolor}
